\pdfoutput=1

\documentclass{article}
\PassOptionsToPackage{dvipsnames,svgnames,table}{xcolor}
\usepackage[preprint]{colm2025_conference}

\usepackage{microtype}
\usepackage{hyperref}
\usepackage{url}
\usepackage{booktabs}
\usepackage[table]{xcolor}
\usepackage{lineno}

\definecolor{darkblue}{rgb}{0, 0, 0.5}
\hypersetup{colorlinks=true, citecolor=darkblue, linkcolor=darkblue, urlcolor=darkblue}

\usepackage{multirow}
\usepackage{latexsym}
\usepackage{enumitem}
\usepackage[T1]{fontenc}
\usepackage{listings}
\usepackage[utf8]{inputenc}

\usepackage{microtype}

\usepackage{inconsolata}

\usepackage{graphicx}
\usepackage{multirow}
\usepackage{booktabs}
\usepackage{wrapfig} 
\usepackage{lipsum} 
\NewDocumentCommand{\heng}
{ mO{} }{\textcolor{red}{\textsuperscript{\textit{Heng}}\textsf{\textbf{\small[#1]}}}}
\NewDocumentCommand{\shujin}
{ mO{} }{\textcolor{cyan}{\textsuperscript{\textit{shujin}}\textsf{\textbf{\small[#1]}}}}
\NewDocumentCommand{\yi}
{ mO{} }{\textcolor{blue}{\textsuperscript{\textit{May}}\textsf{\textbf{\small[#1]}}}}
\NewDocumentCommand{\cheng}
{ mO{} }{\textcolor{orange}{\textsuperscript{\textit{Cheng}}\textsf{\textbf{\small[#1]}}}}
\usepackage{enumitem}
\usepackage{xspace}
\newcommand{\method}{\emph{Alice}\xspace}

\usepackage{graphicx}
\newcommand*{\img}[1]{%
    \raisebox{-.2\baselineskip}{%
        \includegraphics[
        height=\baselineskip,
        width=\baselineskip,
        keepaspectratio,
        ]{#1}%
    }%
}
\title{\img{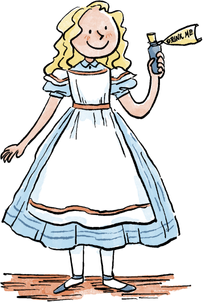} \method: Proactive Learning with Teacher's Demonstrations \\for Weak-to-Strong Generalization}

\author{Shujin Wu$^{1, 2\thanks{Work was done while Shujin Wu was an intern at the University of Illinois Urbana-Champaign.}}$ ~~~Cheng Qian$^{1}$ ~~~ Yi R. (May) Fung$^{1}$ ~~~Paul Pu Liang$^{3}$ ~~~\textbf{Heng Ji}$^{1}$\\
$^{1}$University of Illinois Urbana-Champaign ~~~~~~~~\\
$^{2}$University of Southern California ~~~~~~~~ \\
$^{3}$Massachusetts Institute of Technology ~~~~~~~~ \\
\texttt{\{shujinwu, chengq9, yifung2, hengji\}@illinois.edu}  ~~ \texttt{ppliang@mit.edu} 
}  

\begin{document}
\ifcolmsubmission
\linenumbers
\fi

\maketitle
\begin{abstract}



The growing capabilities of large language models (LLMs) present a key challenge of maintaining effective human oversight.
Weak-to-strong generalization (W2SG) offers a promising framework for supervising increasingly capable LLMs using weaker ones.
Traditional W2SG methods rely on passive learning, where a weak teacher provides noisy demonstrations to train a strong student.
This hinders students from employing their knowledge during training and reaching their full potential.
%
In this work, we introduce \textbf{\method} (pro\textbf{A}ctive \textbf{l}earning w\textbf{i}th tea\textbf{c}her's D\textbf{e}monstrations), a framework that leverages complementary knowledge between teacher and student to enhance the learning process.
We probe the knowledge base of the teacher model by eliciting their uncertainty, and then use these insights together with teachers' responses as demonstrations to guide student models in self-generating improved responses for supervision.
In addition, for situations with significant capability gaps between teacher and student models, we introduce cascade \method, which employs a hierarchical training approach where weak teachers initially supervise intermediate models, who then guide stronger models in sequence. 
Experimental results demonstrate that our method significantly enhances the W2SG performance, yielding substantial improvements in three key tasks compared to the original W2SG: knowledge-based reasoning (+4.0\%), mathematical reasoning (+22.62\%), and logical reasoning (+12.11\%). 
This highlights the effectiveness of our new W2SG paradigm that enables more robust knowledge transfer and supervision outcome. The code is made public at \url{https://github.com/ShujinWu-0814/Alice}.


%

\end{abstract}
\section{Introduction}
%

\looseness=-1
Large Language Models (LLMs) have demonstrated significant capabilities on various tasks~\citep{brown2020language, dubey2024llama, jiang2023mistral}. 
Current evidence suggests LLMs may achieve superior performance compared to humans across many applications~\citep{silver2017mastering, achiam2023gpt, wu2024aligning}.
The rapid progress raises a critical research question: How to provide meaningful supervision on LLMs that surpass human abilities and continually improve their performance~\citep{huang2024superalignment, kim2024road}? 
%


Weak-to-strong generalization (W2SG) tackles this challenge by studying how less capable teacher models (proxy of humans) can supervise more advanced student models~\citep{burns2023weak}. 
The results in~\citet{burns2023weak} reveal that when directly trained on noisy demonstrations (\textit{i.e.,} flawed or incomplete labels) generated by the weak teacher, strong student models can still generalize beyond their teachers' capabilities.
However, the existing W2SG approaches follow a passive learning paradigm, where training solely on noisy responses from weak teachers prevents the students from exploiting their strong capabilities to optimize learning and reach their full potential. 

\begin{figure*}
    \centering
    \includegraphics[width=\linewidth]{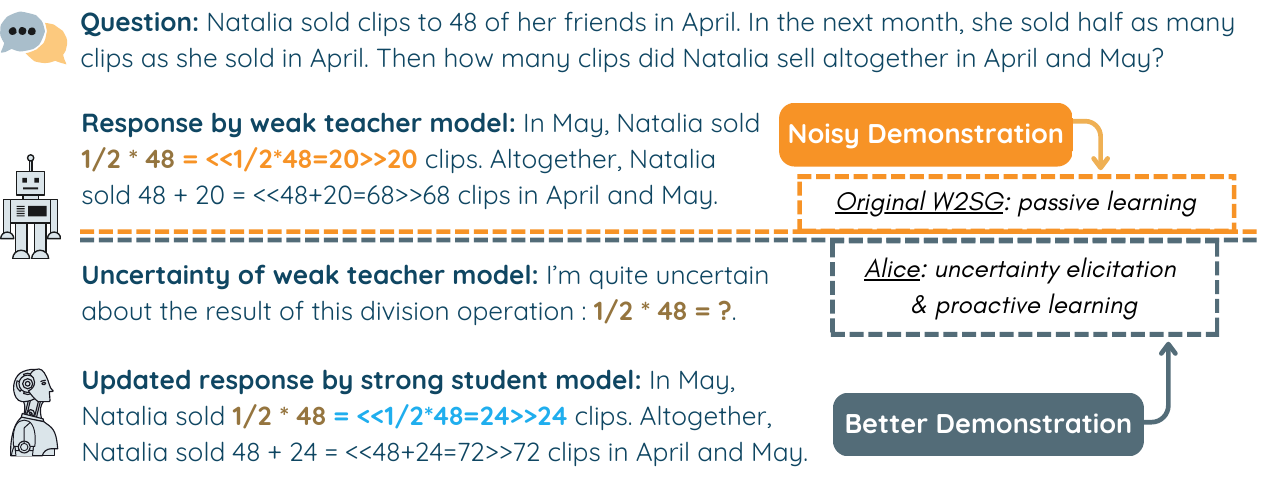}
    \caption{The comparison between the typical W2SG approach and \method.  
    While the typical W2SG approach utilizes noisy demonstrations that may contain misleading information to supervise the strong student directly, we probe the knowledge base of the teacher model and take advantage of the strong student model's capabilities to bridge the knowledge gap and generate higher-quality demonstrations for supervision.}
    \label{fig:enter-label}
\end{figure*}

In this work, we present \textbf{\method} (pro\textbf{A}ctive \textbf{l}earning w\textbf{i}th tea\textbf{c}her's D\textbf{e}monstrations), a paradigm where strong student models are incentivized to generate and refine their own training data to elicit their capabilities, rather than learning passively~\citep{chungincentivize}.
As shown in Figure~\ref{fig:enter-label},
\method starts by probing the real knowledge base of weak teacher models through uncertainty expression~\citep{liu2024can, xu2024sayself} (see ``Uncertainty of weak teacher model'' in Figure~\ref{fig:enter-label}).
%
In the generalization phase, we extend beyond typical student-teacher fine-tuning approaches. Rather than having the student model learn directly from teacher-generated labels, we provide it with three inputs: the teacher model's answer, teacher's uncertainty expression, and the student's own zero-shot response to the question. 
By synthesizing these inputs via zero-shot inference, the student model can leverage both the teacher's task-specific guidance and its superior capabilities to self-generate higher-quality responses to serve as training supervision. 

%
In addition, for scenarios with substantial capability gaps between teacher and student models, we introduce cascade \method, a multi-stage weak-to-strong supervision framework.
Cascade \method builds on the observation that intermediate models trained via \method can even outperform those trained directly on ground-truth labels.
Our approach implements an iterative process where weak teachers first guide intermediate models, which then serve as teachers for stronger models. 
This cascade approach breaks down large capability gaps into manageable steps, enabling more stable knowledge transfer while preserving and enriching the knowledge through each successive stage.

To evaluate the effectiveness of \method, we conduct comprehensive experiments using Qwen-2.5~\citep{yang2024qwen2} and Llama 3.1~\citep{dubey2024llama} model families, each consisting of teacher-student pairs of varying sizes. 
We test these pairs across four datasets that evaluate different capabilities: knowledge-based reasoning, mathematical reasoning, and logical reasoning.
Our experimental results demonstrate significant improvements in supervision performance compared to previous W2SG approaches, with relative gains of 4.0\%, 22.62\%, and 12.11\% across the respective tasks.

\section{Related Work}
A critical research challenge today is developing effective supervision methods for LLMs that exceed human performance, particularly when relying on annotations from average human evaluators. Two complementary approaches addressing this challenge are scalable oversight and weak-to-strong generalization techniques~\citep{jan2024}.

\subsection{Scalable Oversight}
Scalable oversight approaches represent a significant advancement in the supervision of LLMs~\citep{bowman2022measuring}, extending beyond traditional learning from human preference strategies~\citep{christiano2017deep, kaufmann2023survey}. These approaches aim to enhance the effectiveness of human annotators in supervising increasingly sophisticated LLMs by providing them with additional tools and frameworks for evaluation.
Several key methodologies have emerged in this field:
(1) AI Debate Frameworks~\citep{irving2018ai, arnesen2024training, michael2023debate, liang2023encouraging, du2023improving,unleashing2024}: These approaches facilitate structured debates between models to surface relevant evidence and reasoning, making it easier for human annotators to evaluate model outputs. The debate process helps expose underlying assumptions, potential flaws, and alternative viewpoints that might not be immediately apparent to human supervisors.
(2) Critique Model Development~\citep{saunders2022self, mcaleese2024llm}: This methodology focuses on training specialized models to generate detailed analytical feedback that serves as a reference point for human annotators. These critique models can highlight potential issues, inconsistencies, or areas requiring closer examination, effectively augmenting human evaluation capabilities.
(3) Task Decomposition Strategies~\citep{christiano2017deep, wu2021recursively}: Complex supervision tasks are systematically broken down into smaller, more manageable components. This hierarchical approach allows for more focused and accurate human oversight of each subtask, while maintaining coherence in the overall evaluation process.
(4) AI-Assisted Feedback Systems~\citep{bai2022constitutional, lee2023rlaif}: This approach leverages more capable models to provide supervision for other LLMs, creating a hierarchical oversight structure. This method can help standardize evaluation criteria and potentially reduce the evaluation load on human supervisors.
(5) Recursive Reward Modeling~\citep{leike2018scalable}: This iterative approach progressively enhances human supervision capabilities by incorporating increasingly sophisticated models into the evaluation loop. Each iteration builds upon previous insights, creating a more refined and effective oversight process.
These approaches can be used in complement with our method that further calibrate the signals generated by the weak teacher models.

\subsection{Weak-to-Strong Generalization}
\looseness=-1
Weak-to-Strong Generalization approaches leverage advanced algorithms to enable strong student models to learn effectively from noisy demonstrations produced by less capable teacher models~\citep{burns2023weak}. This framework has seen several key developments and applications across different domains.
Recent research has enhanced W2SG through various innovations. 
For instance, ensemble learning techniques have been successfully applied to improve the robustness and effectiveness of W2SG methods~\citep{sang2024improving}. 
\citet{zheng2024weak} adopt weak-to-strong extrapolation to enhance LLMs alignment.
Additionally, the concept of easy-to-hard generalization has emerged as a promising variant of W2SG, where models are initially trained on easily verifiable examples before tackling more complex tasks~\citep{hase2024unreasonable}.
One notable implementation of this approach involves training a strong reward model on human-verifiable examples, which then guide the supervision of more capable models on challenging tasks~\citep{sun2024easy}.
In addition, the effectiveness of W2SG extends beyond LLMs, with successful applications demonstrated in computer vision tasks as well~\citep{guo2024vision}.
In this work, we extend beyond the traditional passive learning paradigm in W2SG by leveraging teacher guidance to calibrate responses from more advanced student models, which offer greater potential.

\section{Method}
\begin{figure*}[t!]
    \centering
    \includegraphics[width=\linewidth]{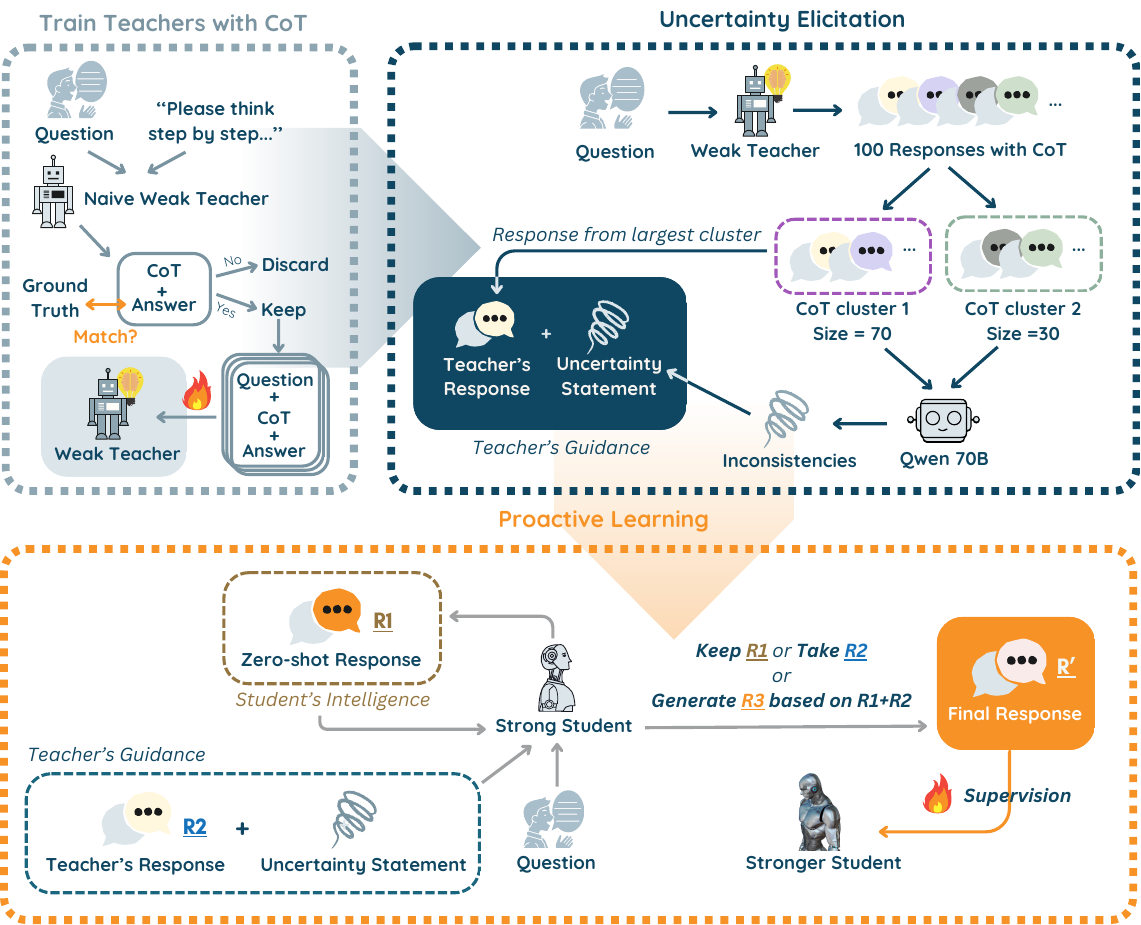}
    \caption{The overview of \method. We first train weak teachers using self-generated CoT to provide them with task-specific knowledge. Next, we probe the teacher models’ knowledge base by eliciting their uncertainty for each question. Finally, we implement proactive learning, where the student model combines teacher guidance with its existing knowledge base and reasoning capabilities to generate final responses.
    These responses are subsequently used to supervise and refine the student model itself.}
    \vspace{-13pt}
    \label{fig:method}
\end{figure*}
We propose \method, a training framework that transforms the typical W2SG solutions from passive to proactive learning. 
Rather than directly fitting the strong students on noisy demonstrations produced by weak teachers, 
we harness the advanced capabilities of student models to self-generate higher-quality responses for supervision with awareness of teachers' demonstrations and uncertainty.


\method consists of two sequential stages following the setting in~\citet{burns2023weak}. 
In the initial stage, we leverage supervised training samples to fine-tune teacher models, enabling them to acquire domain-specific expertise (serving as proxy of humans).
In the second stage, we use unlabeled questions to generate supervision for training student models, simulating the real W2SG scenarios.
We first generate the teacher model's demonstrations, which include both responses and uncertainty expressions. 
Subsequently, student models are guided to incorporate their own solutions to produce refined demonstrations for supervision.
%

\subsection{Fine-Tuning Teacher Models}
\looseness=-1
To prepare the teacher models, we first fine-tune them using supervised examples to establish task-specific knowledge. 
To further improve performance and enable uncertainty expression, we curate the dataset to instruct the models to generate both step-by-step reasoning and final answers for each question. 
For datasets like GSM8K~\citep{cobbe2021training} that include annotated reasoning chains, we directly utilize these existing annotations for fine-tuning. 
For datasets containing only answer labels without chain-of-thought (CoT) annotations, 
we implement the rejection sampling following~\citet{zelikman2022star}: First, we prompt the base teacher models to generate reasoning chains for questions in a zero-shot manner. 
We then validate these generated chains by verifying that their final answers match the annotated ones. Only reasoning chains that produce correct answers are retained for the final teacher model fine-tuning.


\subsection{Eliciting Uncertainty from Teachers}
\looseness=-1
To analyze teacher models' inherent knowledge and uncertainty, we employ a systematic multi-step process based on~\citet{xu2024sayself}.
For each question in the latter half of each dataset's training set, we generate 100 reasoning chains and answers from the teacher models. 
We then use Instructor~\citep{INSTRUCTOR}, an instruction-finetuned text embedding model, to create task-specific and domain-appropriate embeddings for each response. 
To select the representative responses, we perform semantic clustering on the response embeddings through an iterative process~\citep{rokach2005clustering}. 
The clustering begins by selecting an initial response as a reference point and computing cosine similarity scores between its embedding and all other responses. Responses with similarity scores below a threshold T are grouped with the reference response. This process repeats, with new reference points selected from the remaining ungrouped responses, until all responses are assigned to clusters. 
we then randomly select one response from each cluster as its representative. The representative from the largest cluster is designated as the teacher model's final response for the question, as its high frequency suggests it would be the most likely output in single-inference scenarios.

To quantify the teacher models' uncertainty in natural language, we prompt Qwen2.5-70B-Instruct~\citep{yang2024qwen2} to analyze the representative responses from all clusters, focusing on identifying inconsistencies in their reasoning processes. The model then synthesizes these observations into a comprehensive \textit{uncertainty expression} that articulates the specific areas and nature of the teacher models' uncertainty for each question.

\subsection{Proactive Learning}
\looseness=-1
\method effectively leverages both the teacher models' responses and their associated uncertainty expression to generate better demonstrations.
The key innovation is to enable strong student models to actively guide the supervision process.
The process begins with zero-shot inference on the unlabeled question set by the student models.
We then provide each student model with its initial responses alongside the teacher models' outputs and uncertainty expression.
Student models are instructed to analyze the input question thoroughly, then either retain their initial responses or produce improved versions by integrating insights derived from teachers' demonstrations.
This process is conducted via zero-shot inference, and enables complementary knowledge infusion, as the final outputs incorporate information from both the teacher and student models. 
Finally, we use these higher-quality demonstrations to fine-tune the student models, completing the supervision cycle.

\subsection{Cascade Generalization}
For situations where significant capability gaps exist between teacher and student models (specifically, when there are significant disparities in model size), 
we introduce \textbf{cascade \method}, a multi-stage supervision framework that enables progressive knowledge transfer~\citep{soviany2022curriculum, bengio2009curriculum}:
weaker teachers first guide intermediate models, which then serve as teachers for more capable models in an iterative process.
This approach builds on a key insight: 
\textit{intermediate models trained via} \method \textit{can even outperform those trained directly on ground-truth labels}.
By leveraging these enhanced intermediate models as teachers, rather than relying solely on the original weaker teachers, we can more effectively supervise increasingly capable students. 
Cascade \method enhances capability transfer by breaking down large capability gaps into a series of smaller teacher-student transitions. 
By maintaining manageable capability gaps between successive pairs in the sequence, this approach enables robust knowledge transfer.

\section{Experiment}
\subsection{Experiment Settings}

\paragraph{Datasets}
We evaluate our approaches on four datasets across three distinct tasks: knowledge-based reasoning, mathematical reasoning, and logical reasoning. 
For knowledge-based reasoning, we employ \textbf{HotpotQA}~\citep{yang2018hotpotqa}, which features explainable multi-hop question-answer pairs, and \textbf{TriviaQA}~\citep{joshi2017triviaqa}, a challenging reading comprehension dataset. 
For mathematical reasoning, we utilize \textbf{GSM8K}~\citep{cobbe2021training}, comprising linguistically diverse grade-school math word problems that demand multi-step solutions. 
Finally, for logical reasoning, we choose \textbf{ARC (Challenge Set)}~\citep{clark2018think}, which consists of grade-school level multiple-choice science questions. 
For each dataset, we follow the experimental setting in~\cite{burns2023weak},  using the first 50\% of supervised samples for training teacher models and the remaining 50\% of unlabeled questions for the W2SG process.

\paragraph{Models}
To demonstrate the broad applicability of our method, we evaluate it across two distinct model families: \textbf{Qwen 2.5-Instruct}~\citep{yang2024qwen2} and \textbf{Llama 3.1-Instruct}~\citep{dubey2024llama}.
Within each family, we construct two teacher-student pairs of varying model sizes. For Qwen 2.5, we establish a direct generalization relationship between the 1.5B and 3B models, and a cascade generalization setup where the 1.5B model teaches the 7B model through an intermediate 3B teacher. 
Similarly, for Llama 3.1, we evaluate the approaches in both direct teaching (1B to 3B) and cascade generalization (1B to 8B through intermediate 3B) settings. 
Since Llama 3.1 doesn't include an 8B variant, we utilize the corresponding Llama 3.2 model instead for this configuration.

\begin{figure*}[t!]
    \centering
    \includegraphics[width=\linewidth]{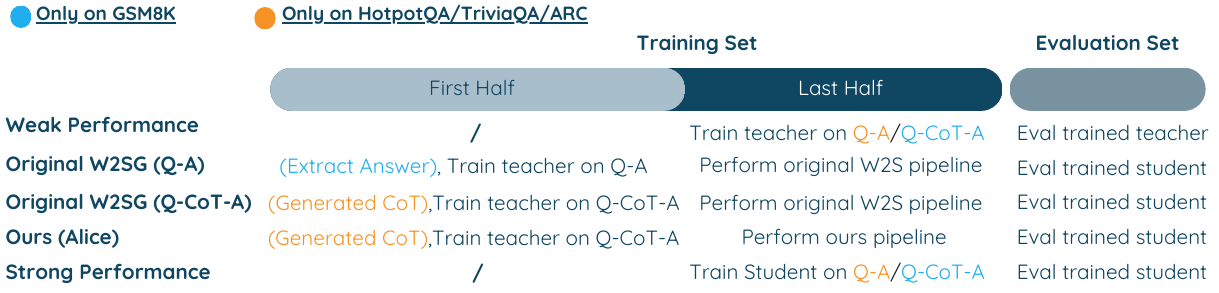}
    \caption{\looseness=-1 Our main experimental settings. For Weak/Strong Performance, we directly fine-tune models on ground-truth labels using the last half of training set before evaluation. For both original W2SG and \method, we first fine-tune teacher models on ground-truth labels using the first half of training set to to equip them with basic task-relevant knowledge, then perform corresponding student supervision using only questions from last half of the training set.}
    
    \label{fig:setting}
\end{figure*}

\paragraph{Baselines}
We include the following baselines for comparison (the implementations are visualized in Figure \ref{fig:setting}): 
\textbf{(1) Weak Performance}: 
We directly fine-tune weak teacher models on the golden labels (Q-CoT-A pairs for GSM8K and Q-A pairs for the rest of three datasets) using the latter half of training set and evaluate them on the evaluation set. 
\textbf{(2) Original W2SG}~\citep{burns2023weak}: 
While the original W2SG method uses teacher-generated answers solely as supervision labels for student models, \method extends this by incorporating CoT reasoning. 
This enhancement leads us to investigate whether the format of teacher-generated labels impacts the original W2SG's performance.
We consider two variants of the original W2SG:
\textbf{a) Original W2SG (Q-A)}: We initially fine-tune weak teacher models using Q-A pairs from the first half of the training set. For the GSM8K dataset, although the golden answer labels include step-by-step reasoning chains, we extract only the final numerical values for fine-tuning.
\textbf{b) Original W2SG (Q-CoT-A)}: We enhance the training by incorporating automatically generated CoTs~\citep{zelikman2022star} alongside answers, utilizing Q-CoT-A pairs from the first half of the training set. 
For GSM8K, we leverage the human-annotated CoT.
In both variants, we first conduct zero-shot inference using teacher models on questions from the second half of the training dataset. We then train student models directly on these teacher-generated demonstrations and evaluate their performance on the evaluation set.
\textbf{(3) Strong Performance}: We directly fine-tune strong student models on the golden labels using the latter half of training set and evaluate them on the evaluation set.

\looseness=-1
\paragraph{Metrics} We evaluate student models' performance using accuracy on the evaluation set. 
For the GSM8K dataset, correctness is determined by an exact match between the model's output and the ground truth. For HotpotQA, TriviaQA, and ARC-Challenge, we employ a verification approach, requiring the model's response to contain the correct answer.

\subsection{Experiment Results}

\label{results}
\looseness=-1
The experimental results are presented in Table \ref{tab:results}:
\begin{itemize} [topsep=1pt, partopsep=1pt, leftmargin=12pt, itemsep=1pt]
\item \textbf{Impact of CoT Integration in W2SG}: Our experimental results demonstrate that incorporating CoT reasoning into the W2SG framework's generalization process (Q-CoT-A) consistently outperforms the traditional question-answer (Q-A) baseline. These findings suggest that training models to employ structured reasoning processes is more effective than direct answer generation. This empirical validation supports our design decision to incorporate CoT reasoning in \method for student model supervision.

\item \textbf{The superior performance of \method}: Compared to the baseline W2SG approach, \method demonstrates substantial performance gains across all four datasets and teacher-student configurations. 
The improvements are consistently notable in three key areas: knowledge-based reasoning (+4.0\%), mathematical reasoning (+22.62\%), and logical reasoning (+12.11\%). 

\item \looseness=-1 \textbf{Notable Improvement on mathematical reasoning}: \method shows particularly strong performance on mathematical reasoning tasks, suggesting that effectively leveraging student model capabilities for proactive learning is crucial for addressing complex problem domains where teachers can only provide limited supervision.

\item \textbf{Proactive learning outperforms fine-tuning on ground-truth labels}: While the conventional W2SG method only partially bridges the performance gap between weak teachers and strong students—resulting in generalization performance intermediate between weak and ground-truth performance—\method enables models to consistently exceed the model performance when trained on ground-truth labels. This establishes the foundation for cascade W2SG, where progressively improved intermediate models provide continuous supervision for stronger models.

\item \textbf{\method combines the teachers' expertise with students' strong capabilities}: \method allows the strong student models to refine their responses while being aware of teacher demonstrations.
Please refer to a case study in Appendix~\ref{sec:appendix} that justifies this statement.

\end{itemize}
 
\begin{table*}[t!]
\centering
\resizebox{\textwidth}{!}{
\def\arraystretch{1.15}
\begin{tabular}{cc|cccc|cccc}
\specialrule{2pt}{1pt}{4pt}
\multicolumn{2}{c|}{}                                                           & \multicolumn{4}{c|}{\textbf{Qwen 2.5 1.5B to 3B}}                              & \multicolumn{4}{c}{\textbf{Qwen 2.5 1B to 7B (\textit{cascade})}}                                 \\ \cmidrule(rl){3-6} \cmidrule(rl){7-10}  
\multicolumn{2}{c|}{}                                                           & \multicolumn{2}{c}{\textbf{Knowledge}} & \textbf{Math} & \textbf{Reasoning} & \multicolumn{2}{c}{\textbf{Knowledge}} & \textbf{Math} & \textbf{Reasoning} \\
\multicolumn{2}{c|}{}                                                           & HotpotQA             & TriviaQA           & GSM8K         & ARC-Challenge      & HotpotQA            & TriviaQA            & GSM8K         & ARC-Challenge      \\ \midrule
\multicolumn{2}{c|}{\textit{Weak}}                                              & 9.32               & 32.96            & 48.18       & 73.98            & 9.32              & 32.96             & 48.18       & 73.98            \\
\multicolumn{1}{c|}{}                                        & \textit{Q-A}     & 11.59              & 30.47            & 12.21       & 70.56            & 16.38             & 37.53             & 14.18       & 74.23            \\
\multicolumn{1}{c|}{\multirow{-2}{*}{\textit{Original W2S}}} & \textit{Q-CoT-A} & 22.21              & 45.76            & 57.71       & 76.37            & 24.29             & 48.95             & 64.39       & 77.26            \\
\rowcolor[HTML]{DBE5EA} 
\multicolumn{2}{c|}{\cellcolor[HTML]{DBE5EA}\textit{\method}}                      & 22.98              & 51.69            & 72.27       & 80.95            & 28.83             & 64.08             & 79.16       & 90.09            \\
\multicolumn{2}{c|}{\textit{Strong}}                                           & 14.76              & 42.45            & 61.26       & 79.10            & 22.38             & 53.44             & 70.99       & 89.33            \\ \specialrule{2pt}{3pt}{4pt}
                                                             &                  & \multicolumn{4}{c|}{\textbf{Llama 3.2 1B to 3B}}                               & \multicolumn{4}{c}{\textbf{Llama 3.2 1B to Llama 3.1 8B {(\textit{cascade})}}}                      \\ \cmidrule(rl){3-6} \cmidrule(rl){7-10}
                                                             &                  & \multicolumn{2}{c}{\textbf{Knowledge}} & \textbf{Math} & \textbf{Reasoning} & \multicolumn{2}{c}{\textbf{Knowledge}} & \textbf{Math} & \textbf{Reasoning} \\
\multicolumn{2}{c|}{}                                                           & HotpotQA             & TriviaQA           & GSM8K         & ARC-Challenge      & HotpotQA            & TriviaQA            & GSM8K         & ARC-Challenge      \\ \midrule
\multicolumn{2}{c|}{\textit{Weak}}                                              & 15.84              & 33.04            & 29.10        & 53.84            & 15.84             & 33.04             & 29.10        & 53.84            \\
\multicolumn{1}{c|}{}                                        & \textit{Q-A}     & 16.81                     & 37.40                    & 5.69        & 57.25                   & 18.57                    & 46.54                    & 5.38              & 46.76                   \\
\multicolumn{1}{c|}{\multirow{-2}{*}{\textit{Original W2S}}} & \textit{Q-CoT-A}          & 25.56            & 57.76                   & 40.03       & 61.89                   & 26.64                    & 62.38                    & 45.64              & 64.11                   \\
\rowcolor[HTML]{DBE5EA} 
\multicolumn{2}{c|}{\cellcolor[HTML]{DBE5EA}\textit{\method}}                      & 22.09                     & 59.91                   & 68.91              & 75.02                   & 27.08                    & 68.91                    & 77.91              & 81.95                  \\
\multicolumn{2}{c|}{\textit{Strong}}                                           & 19.05              & 52.18            & 54.03        & 72.61            & 25.20                    & 63.74                    & 64.22              & 74.58                   \\ \bottomrule[2pt]
\end{tabular}}
\caption{\label{tab:results}Main experimental results across four datasets and four teacher-student model pairs. \method significantly outperforms the original W2SG method and even surpasses the strong performance at most times.}
\end{table*}


\section{Further Analysis}

\subsection{Ablation Study}
\looseness=-1
To evaluate our design choices, we conduct an ablation study using Qwen 2.5 models, with the 1.5B variant as the teacher and the 7B variant as the student. We compare \method against two ablated variants:
(1) \textbf{w/o Cascade}: Removing the cascade generalization process, instead having the 1.5B model directly teach the 7B model using \method.
(2) \textbf{w/o Teacher's Uncertainty}: Removing the teacher's knowledge base probing through uncertainty elicitation during cascade stages, while maintaining the teacher's responses during proactive learning.

The results, presented in Table \ref{tab:Ablation}, reveal two key findings. 
First, removing cascade generalization significantly degrades performance across all datasets, with the most substantial decrease (4.65\%) observed on ARC-Challenge. This demonstrates that breaking down the learning process into incremental steps through cascade generalization enables more effective knowledge transfer and improved generalization performance. 
Second, removing the teacher's uncertainty consistently reduces performance, indicating that probing the teacher model's knowledge base and incorporating its uncertainty statements are crucial for generating high-quality demonstrations and achieving better supervision outcomes.

\subsection{Effect of Multiple Teachers}

We conduct experiments to understand the effect of including extra teacher models in the W2SG process. 
First, we investigate the effectiveness of incorporating multiple teacher models in the generalization process by simultaneously providing uncertainty expression and responses from both Qwen 1.5B and 3B models to the 7B model without performing cascade generalization.
The 7B model is then tasked with generating final answers based on the combined guidance. 
As shown in Table \ref{tab:Ablation}, this Multi-teacher approach demonstrates consistent improvements over the baseline without cascade training (\textit{w/o Cascade}), which uses only the 1.5B model as the teacher. 
Specifically, utilizing multiple teacher models yields an average increase of 2.22\% in test accuracy across all four datasets, demonstrating the value of additional teacher supervision.

We further compare the Multi-teacher approach with cascade \method, as both utilize the 3B model but in fundamentally different ways. In Multi-teacher, a trained 3B model serves as a concurrent teacher alongside the 1.5B model, while in cascade \method, an untrained 3B model functions as an intermediate step in a sequential knowledge transfer process to the 7B model. 
Our experiments show that cascade \method consistently achieves superior performance across all datasets. 
This suggests that while multiple trained teachers can improve performance, the cascade generalization strategy is more effective due to its structured, sequential approach to knowledge integration across model scales. 
We hypothesize that simultaneous guidance from two trained teachers (1.5B and 3B) may introduce interference in the learning process, making it challenging for the 7B model to optimally integrate information from multiple concurrent sources.

\begin{table*}[t!]
\centering
\resizebox{\textwidth}{!}{
\def\arraystretch{1.2}
\begin{tabular}{cccccc}
\specialrule{2pt}{1pt}{4pt}
\multicolumn{1}{l}{}                & \multicolumn{1}{l|}{}                                                                            & \multicolumn{2}{c}{\textbf{Knowledge}} & \textbf{Math} & \textbf{Reasoning} \\
\textbf{Method}                     & \multicolumn{1}{c|}{\textbf{Setting}}                                                            & HotpotQA            & TriviaQA            & GSM8K         & ARC-Challenge      \\ \midrule
\rowcolor[HTML]{FCD190} 
\textit{\method}                       & \multicolumn{1}{c|}{\cellcolor[HTML]{FCD190}\textit{Qwen 1.5B -\textgreater 3B -\textgreater 7B}} & 28.83             & 64.08             & 79.16       & 90.09            \\
\textit{w/o Cascade} & \multicolumn{1}{c|}{\textit{Qwen 1.5B -\textgreater 7B}}                                         & 27.68             & 60.48             & 76.78       & 85.44            \\
\textit{w/o Teacher's Uncertainty}  & \multicolumn{1}{c|}{\textit{Qwen 1.5B -\textgreater 3B -\textgreater 7B}}                                         & 25.86             & 63.04             & 72.66       & 88.96            \\
\textit{Multi-teacher}               & \multicolumn{1}{c|}{\textit{Qwen 1.5B+ 3B -\textgreater 7B}}                                     & 28.25             & 63.54             & 77.61       & 89.86            \\
\textit{Mix-teacher}                 & \multicolumn{1}{c|}{\textit{Qwen 1.5B + Llama 1B -\textgreater Qwen 7B}}                         &28.17                     &63.75                     &77.47              &90.48                    \\
\textit{Cross-model}                 & \multicolumn{1}{c|}{\textit{Llama 1B -\textgreater Qwen 7B}}                                     &28.16                     &62.53                     &70.66               &89.92                    \\
\textit{Cross-model-Cascade}         & \multicolumn{1}{c|}{\textit{Qwen 1.5B -\textgreater Llama 3B -\textgreater Qwen 7B}}             &28.91                     &64.92                     &79.47               &90.38                    \\ \bottomrule[2pt]
\end{tabular}}
\caption{\label{tab:Ablation} Results for ablation studies and further analysis.}
\vspace{-5pt}
\end{table*}

\subsection{Cross-Family Knowledge Transfer Effects}
We conduct experiments to understand the effect of infusing knowledge from different model families in W2SG. 
First, we use a Llama 3 1B model to teach a Qwen 7B model (referred to as \textit{Cross-model}). 
In Table \ref{tab:Ablation}, compared to our baseline without cascading (\textit{w/o cascade}), \textit{Cross-model} shows improved generalization performance across all four datasets when using the cross-family teacher model. 
This suggests that teachers from different model families can provide complementary perspectives and knowledge that enhance the student's learning process.
To further explore this effect, we compare two cascading approaches: \textit{Cross-model Cascade}, which uses Llama 3B as an intermediate model, and cascade \method, which uses Qwen 3B. 
The superior performance of \textit{Cross-model Cascade} provides additional evidence that cross-family knowledge transfer can enhance model performance.
We also examine the impact of combining teachers from different families in a multi-teacher setting. 
We compare the \textit{Multi-teacher} approach, which uses two Qwen models in different sizes as teachers, with \textit{Mix-teacher}, where we replace one Qwen 3B teacher with a Llama 1B model. 
Despite the Llama teacher's smaller size, \textit{Mix-teacher} achieves comparable performance to \textit{Multi-teacher}, supporting the claim that diverse teaching perspectives brought by different model families contribute positively to model generalization.

\subsection{Cascade Generalization on Original W2SG}
\setlength{\intextsep}{0pt}  
\setlength{\columnsep}{15pt}  
\begin{wraptable}{r}{0.6\textwidth}
\centering
\resizebox{\linewidth}{!}{
\def\arraystretch{1.2}
\begin{tabular}{ccccc}
\specialrule{2pt}{1pt}{4pt}
\multicolumn{1}{c|}{}                                          & \multicolumn{2}{c}{\textbf{Knowledge}} & \textbf{Math}    & \textbf{Reasoning} \\
\multicolumn{1}{c|}{\textbf{}}                                 & HotpotQA            & TriviaQA            & GSM8K            & ARC-C      \\ \midrule
\multicolumn{5}{c}{\cellcolor[HTML]{DBE5EA}\textit{\textbf{Original W2SG}}}                                                                        \\ \midrule
\multicolumn{1}{c|}{\textit{3B -\textgreater{}7B (direct)}}    & 28.55             & 61.87             & 69.04          & 84.65            \\
\multicolumn{1}{c|}{\textit{1.5B -\textgreater{}7B (cascade)}} & 26.31             & 57.19             & 67.53          & 83.36            \\ \midrule
\multicolumn{5}{c}{\cellcolor[HTML]{FCD190}\textit{\textbf{\method}}}                                                                                 \\ \midrule
\multicolumn{1}{c|}{\textit{3B -\textgreater{}7B (direct)}}    & 27.59             & 62.56             & 77.98          & 87.11            \\
\multicolumn{1}{c|}{\textit{1.5B -\textgreater 7B (cascade)}}  & \textbf{28.83}    & \textbf{64.08}    & \textbf{79.16} & \textbf{90.09}   \\\bottomrule[2pt]
\end{tabular}}
\caption{\label{tab:original-cascade}The comparison between applying cascade generalization to original W2SG and to \method.}
\vspace{10pt}
\end{wraptable}

We investigate the varying impact of cascade generalization when applied to the original W2SG versus our proposed method.
Our experimental results demonstrate that models trained with \method consistently exceed the strong performance (see section \ref{results}), while those trained with original W2SG do not. 
This observation suggests a key difference in the intermediate stages: when implementing cascade generalization with our approach, the intermediate model—trained by having a 1.5B model teach a 3B model in stage 1—outperforms a standard 3B model. 
This enhanced intermediate model, when used to teach a 7B student in the subsequent stage, produces superior results compared to direct teaching from a standard 3B teacher to a 7B student. 
In contrast, cascade generalization applied to the original W2SG fails to enhance the intermediate teacher's capabilities. The 3B model produced after stage 1 performs comparably to a standard 3B model, resulting in worse final performance after the subsequent stage compared to direct teaching from a 3B to a 7B model.

To validate these findings, we conduct experiments across four datasets under these conditions. 
The results in Table \ref{tab:original-cascade} demonstrate that when comparing direct teaching (3B to 7B) versus cascade teaching (1.5B to 3B to 7B), our method consistently outperforms the baseline, while the original W2SG shows no improvement.
These findings have important implications for scenarios where AI models evolve beyond human-level capabilities. 
\method demonstrates that effective supervision can be maintained using relatively smaller teacher models (1.5B), reducing the need to continuously increase teacher model size. 
The original W2SG approach, in contrast, requires teacher models to scale proportionally with student capabilities—an unsustainable approach given the expected widening capability gap between human supervisors and advanced AI systems.
Our method thus provides a more practical framework for supervising increasingly capable models while maintaining robust oversight with constrained supervision resources.

\section{Conclusion}
We propose \method, a proactive learning paradigm with bidirectional teacher-student interaction.
By probing teacher model uncertainty and leveraging improved student-generated demonstrations, our approach enhances supervision effectiveness. 
Furthermore, cascade \method enables hierarchical training for scenarios with significant capability gaps. 
Experimental results demonstrate that our approach significantly improves W2SG performance, establishing a new promising direction for supervising superhuman models with limited human oversight.
We discuss the limitations in Appendix~\ref{sec:limitation}.




\bibliography{main}
\bibliographystyle{colm2025_conference}

\appendix

\section*{Appendix}

\section{Prompt}
\subsection{Uncertainty Summarization Prompt}
\begin{lstlisting}
Your task is to analyze a question provided to you along with several responses generated by my model. Your objective is to identify and summarize the inconsistency in the models' responses that can explain why my model is uncertain about the correct answer.

Please note that:
1. You should give the reasons from a first-person perspective, as if you are my model that gives the provided responses.
2. Limit your explanation to the knowledge and facts the model possesses about the question.
3. Keep your summary brief, aiming for 1-3 sentences.
4. Please directly provide the summarized reason without any greetings or other unnecessary information. If you find all the responses are quite the same regarding the question, please directly return N/A. 
5. Importantly!! My model only has access to one response at a time. Thus, the summary you provided should not include any statement like "My different responses have...", "my multiple responses about ...", etc. You should not say "my responses" or "the responses" anywhere in the summary. Just simply provide the uncertainty.

Here is an example:
Question: Sky High starred the actress who is married to which actor?

Responses:
1. The actress who starred in "Sky High" (2005) and is married to an actor is Kelly Preston. Her husband is John Travolta. The two have been married since 1991 and have three children together.
2. The actress who starred in "Sky High" (2005) and is married to an actor is Kristen Bell. Bell voiced the main character, Layla, in "Sky High," and she is married to Dax Shepard, who is also an actor.
3. The actress who starred in "Sky High" (2005) and is married to an actor is Kelly Clarkson. Her acting debut was in this film, and she married singer and actor Brandon Blackstock in 2013.

The output can be: I am uncertain about the correct actress in "Sky High". There is a probability that the actress is Kristen Bell, instead of Kelly Preston. I am confused about her voice acting roles with on-screen appearances. There is also some probability that the actress is Kelly Clarkson.

Now consier the following case:
Question: 
{}

Responses: 
{}
\end{lstlisting}

\subsection{Prompt to Student Model}
\begin{lstlisting}
You will be provided with a question, a response from yourself, and a response from another model and the uncertainty statement from that model as well.

You should take another model's response and the uncertainty statement as references to help you provide a final response to the question. You can either keep your original response, believe in another model's response, or generate a new one based on all these given information. You are also required to provide your step-by-step solution to the question.

Here is the question: {}
Here is your original answer: {}
Here is another model's answer: {}
Here is another model's uncertainty statement: {}

Note: You should not mention the model's response or the uncertainty statement in your new response. Just simply output your final response to the given question. Do not explain about why you choose to keep your original response or give a new one.

IMPORTANT: You must provide an answer to the question. You should not say 'I don't know' or 'I am not sure'. You should always try your best to provide an answer at your best knowledge.
\end{lstlisting}

\begin{figure*}[t!]
    \centering
    \includegraphics[width=\linewidth]{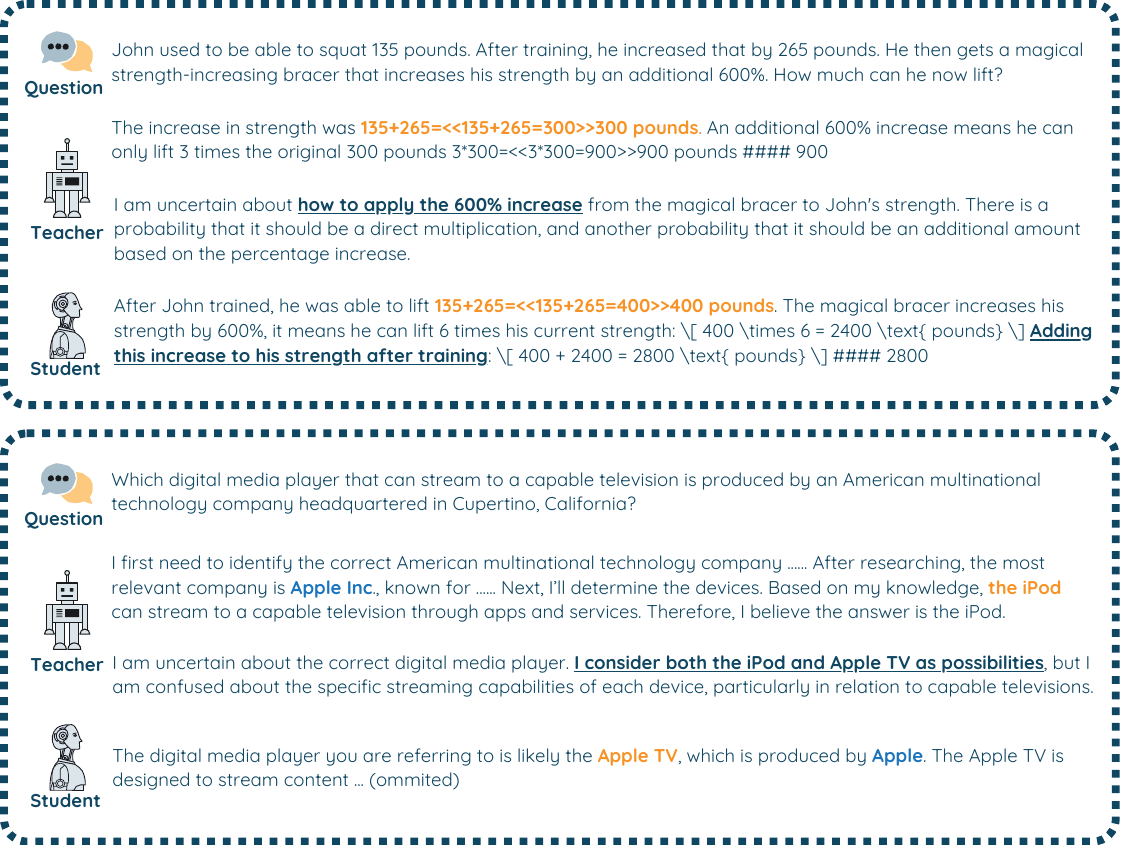}
    \caption{Case studies of \method's effectiveness in generating higher-quality supervision signals by eliciting teacher's uncertainty and taking advantages of student's superior capabilities to bridge the potential knowledge gap.}
    \label{fig:cases}
\end{figure*}

\section{Case Study}
\label{sec:appendix}
We conduct a case study to demonstrate the effectiveness of our approach (see Figure \ref{fig:cases}). 
We examine two representative examples: one from the GSM8K dataset and the other from HotpotQA. 
In the first case, the teacher's uncertainty stems from ambiguity in calculating percentage increases—specifically, whether to apply direct multiplication or add the percentage to the original amount. By analyzing this information alongside the teacher's original response, the student model leverages its enhanced reasoning capabilities to determine that the 600\% increase should be added to the previously calculated amount. Furthermore, it identifies and corrects an arithmetic error in the teacher's calculation [135+265 = 300], demonstrating its ability to catch hidden mistakes even when they are not explicitly captured in the teacher's uncertainty statement. 
In the second case, after confirming the company is Apple, the teacher model expresses uncertainty about which specific product meets the question's requirements. It explicitly acknowledges that either the iPod or the Apple TV could be the answer. Based on this uncertainty, the student model analyzes the streaming capabilities of both products and correctly determines that the Apple TV, not the iPod, satisfies the criteria. 
These cases clearly showcase that \method can effectively probes the teacher model's knowledge base, optimally leverages the distinct capabilities of both teacher and student models, and ultimately produces higher-quality demonstrations for more effective subsequent supervision.

\section{Limitations}
\label{sec:limitation}
While our work demonstrates promising results in enhancing W2SG performance, several important limitations should be noted. 
First, \method relies heavily on how teacher model's uncertainties are articulated. 
In cases where the teacher model's uncertainty expression fails to be accurately elicited, the framework's effectiveness may be significantly reduced, potentially leading to suboptimal supervision results. 
Next, our approach implements only a single-turn learning paradigm, although an ideal setting should enable iterative, dynamic interaction between teacher and student models.  
Such a multi-turn approach would theoretically enable the teacher to express initial uncertainty, receive the student's proposed solution, assess its validity, and provide feedback until reaching a consensus. While our preliminary experiments with a two-turn interaction proved unpromising, leading us to just focus on current single-turn approach, we still believe that exploring more effective mechanisms for multi-turn model interactions remains an important direction for improving generalization performance. Finally, our current cascade \method approach only implements a two-stage cascade process, primarily due to practical constraints. These include the limited availability of models with varying sizes and computational resource restrictions. However, as the capability gap between teacher and student models continues to widen in future scenarios, decomposing the supervision into multiple intermediate stages should be implemented to ensure stable knowledge transfer and reach the optimal supervision outcome.
\end{document}